\documentclass[journal]{IEEEtran}

\usepackage{graphicx}
\usepackage{amsmath}
\usepackage{amssymb}
\usepackage{tabularx}
\usepackage{subcaption}
\usepackage{times}
\usepackage{multicol}
\usepackage{adjustbox}
\usepackage{xcolor}
\usepackage{cleveref}
\usepackage{url}
\usepackage{cite}
\usepackage{placeins} % 用於 \FloatBarrier

\begin{document}

% paper title
\title{EfficienT-HDR: An Efficient Transformer-Based Framework via Multi-Exposure Fusion for HDR Reconstruction}

% author names and IEEE memberships
\author{
    Yu-Shen Huang, Tzu-Han Chen,
    Cheng-Yen Hsiao, and Shaou-Gang Miaou
}

% make the title area
\maketitle

% As a general rule, do not put math, special symbols or citations
% in the abstract or keywords.
\begin{abstract}
Achieving high-quality High Dynamic Range (HDR) imaging on resource-constrained edge devices is a critical challenge in computer vision, as its performance directly impacts downstream tasks such as intelligent surveillance and autonomous driving. Multi-Exposure Fusion (MEF) is a mainstream technique to achieve this goal; however, existing methods generally face the dual bottlenecks of high computational costs and ghosting artifacts, hindering their widespread deployment.

To this end, this study proposes a light-weight Vision Transformer architecture designed explicitly for HDR reconstruction to overcome these limitations. This study is based on the Context-Aware Vision Transformer and begins by converting input images to the YCbCr color space to separate luminance and chrominance information. It then employs an Intersection-Aware Adaptive Fusion (IAAF) module to suppress ghosting effectively. To further achieve a light-weight design, we introduce
Inverted Residual Embedding (IRE), Dynamic Tanh (DyT), and  propose Enhanced Multi-Scale Dilated Convolution (E-MSDC) to reduce computational complexity at multiple levels.

Our study ultimately contributes two model versions: a
main version for high visual quality and a light-weight version
with advantages in computational efficiency, both of which
achieve an excellent balance between performance and image
quality. Experimental results demonstrate that, compared to
the baseline, the main version reduces FLOPS by approximately
67\% and increases inference speed by more than
fivefold on CPU and 2.5 times on an edge device. These results
confirm that our method provides an efficient and ghost-free
HDR imaging solution for edge devices, demonstrating 
versatility and practicality across various dynamic scenarios.
\end{abstract}


\section{Introduction}

In recent years, edge computing has rapidly developed alongside the proliferation of smart devices and the Internet of Things (IoT). One of its core challenges lies in achieving high image quality and processing efficiency under limited computational resources. In applications like intelligent surveillance, autonomous driving, and augmented reality (AR), stable and high-quality imagery plays a critical role in subsequent perception and decision-making tasks. However, in complex real-world lighting conditions, the limited dynamic range of standard sensors cannot simultaneously capture details in  bright highlights and deep shadows. High Dynamic Range (HDR) imaging is the key solution to overcome this limitation, but achieving high-quality HDR on resource-constrained edge devices constitutes a significant challenge.

\begin{figure}[t!]
    \centering
    \includegraphics[width = 0.7\linewidth]{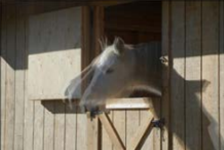}
    \caption{Illustration of ghosting in image fusion~\cite{ma2017}.}
    \label{fig:ghosting}
\end{figure}

Multi-Exposure Fusion (MEF) is a widely adopted computational photography technique used to reconstruct an HDR image by combining multiple low dynamic range (LDR) images taken at different exposures~\cite{liu2023}\!\cite{yang2023}.
 While this approach holds great potential, existing MEF techniques generally face two major bottlenecks when aiming for HDR reconstruction, which hinder their practical deployment on edge devices. First, in dynamic scenes with moving objects, misalignments between exposures lead to severe ghosting artifacts, as shown in Fig.~\ref{fig:ghosting}, which degrade visual quality and reduce the reliability of image analysis~\cite{ma2017}. Second, many state-of-the-art algorithms, especially those based on deep learning, have high computational costs and energy consumption, making them inefficient to run on edge devices.

To address these challenges and enable practical, high-quality HDR imaging on edge devices, this study proposes a novel light-weight Vision Transformer framework named EfficienT-HDR. This framework aims to achieve efficient, ghost-free HDR reconstruction by improving the existing MEF pipeline by integrating several innovative modules designed to tackle ghosting and computational costs.

Our main contributions are as follows:
\begin{itemize}
    \item \textbf{Light-weight Vision Transformer architecture:} We propose a light-weight Vision Transformer architecture that integrates Inverted Residual Embedding (IRE)~\cite{sandler2019}, Dynamic Tanh (DyT) modules~\cite{zhu2025}, and our Enhanced Multi-Scale Dilated Convolution (E-MSDC)~\cite{gao2025}. The design addresses the high computational cost of conventional Vision Transformers and enhances the feasibility of deploying the model on edge devices.
    \item \textbf{Utilize Intersection-Aware Adaptive Fusion (IAAF) module in HDR:} We utilize an Intersection-Aware Adaptive Fusion (IAAF) module~\cite{weng2024 }\!\cite{weng2024rethinking}, which removes redundant information by learning feature intersections. This light-weight design effectively suppresses ghosting artifacts in dynamic scenes, preserves unique details from each exposure, and improves the overall stability and quality of the fused output.
    \item \textbf{High efficiency with low computational cost:} Compared to the original model, floating-point operations (FLOPS) are reduced by approximately 67\%, and the CPU inference speed is increased by more than five times. These results demonstrate that the architecture effectively balances performance and efficiency.
\end{itemize}

\section{Related Work}
This study focuses on addressing three key challenges: exposure correction, multi-exposure fusion, and high dynamic range (HDR) imaging. The following provides a literature review on these three topics.

\subsection{Exposure Correction}
In digital image processing and photography, exposure refers to the total amount of light received by the image sensor, which determines the brightness and clarity of the captured image. Proper exposure adjustment prevents overexposure or underexposure, where the former results in the loss of highlight details, producing pure white regions, and the latter causes shadow details to disappear, yielding entirely black areas. Since a single image cannot simultaneously preserve details in bright and dark areas, especially in high dynamic range scenes, multi-exposure image fusion techniques have emerged to address this limitation.

In traditional methods, Contextual and Variational Contrast Enhancement~\cite{celik2011}, performs exposure correction using histogram-based approaches, while the Retinex Theory~\cite{land1977} achieves similar effects based on the Retinex model. However, these methods often struggle to balance highlight and shadow details under extreme illumination conditions or in complex scenes. Deep learning-based multi-exposure image fusion methods, such as~\cite{ma2020}, achieve more uniform and natural brightness by predicting and adjusting the weighting of exposure regions. However, these methods are computationally intensive and require substantial resources. Against this backdrop, Weng et al.~\cite{weng2024b} proposed an exposure correction approach based on the Atmospheric Scattering Model (ASM) combined with the Retinex theory. The core formulation is written as:
\begin{equation}
I(x) = t(x) \cdot J(x) + \big(1 - t(x)\big) \cdot A
\label{eq:asm}
\end{equation}
where $I(x)$ denotes the observed image, $J(x)$ is the clear image, $t(x)$ represents the transmission map, and $A$ is the atmospheric light. This method performs exposure correction through local and global branches. The local branch leverages prior knowledge to capture features in bright and low-light regions, while the global branch uses gamma correction to optimize brightness distribution further. This approach produces more natural visual results and outperforms existing methods in image quality assessment, demonstrating the effectiveness of combining dark and bright channel priors with gamma correction to enhance image quality and detail representation. This study will focus on applying prior knowledge and exploiting the advantages of differently exposed images to improve overall image quality effectively.

\subsection{Multi-Exposure Fusion}
Non-deep learning MEF methods typically rely on classical image processing algorithms, which can be broadly categorized into spatial-domain and transform-domain approaches. Spatial-domain methods focus on local detail enhancement. For example, Ma and Wang~\cite{ma2015} proposed a patch-based fusion method that processes images region by region, effectively preserving image details and mitigating exposure inconsistencies. Li et al.~\cite{li2021} introduced a technique that combines detail-preserving factors with adjustable weighting curves, which corrects edge detail loss and balances bright and dark regions, thus improving both image quality and computational efficiency. Transform-domain methods, on the other hand, process images by converting them into the frequency or multi-scale domain. For instance, the gradient pyramid model proposed by~\cite{burt1993} laid the theoretical foundation for early MEF methods, while Mertens et al.~\cite{mertens2007} employed a Laplacian pyramid structure for multi-scale fusion. By computing weights based on contrast, saturation, and exposure, these methods generate high-quality images and have been widely applied to high dynamic range (HDR) imaging.

With the rapid development of deep learning, methods based on the convolutional neural network (CNN) have emerged. Several MEF-Nets were proposed to predict low-resolution weight maps and combines them with guided filters to achieve high-resolution fusion, significantly reducing the computational load. DPE-MEF~\cite{han2022} employs a detail enhancement module and a color enhancement module to preserve details and optimize color simultaneously. DeepFuse~\cite{prabhakar2017}, an unsupervised approach, achieves efficient learning by fusing low-level features. TransMEF~\cite{qu2021} utilizes an encoder-decoder architecture to learn multi-exposure characteristics in a self-supervised training framework. However, most of these methods mainly target static scenes and are still limited in handling ghosting artifacts in dynamic scenarios.

Generative adversarial network (GAN) based methods, such as MEF-GAN~\cite{xu2020}, leverage a generator to learn multi-exposure features and a discriminator to enhance the realism of the fused images. Although GAN-based approaches demonstrate potential for producing high-quality images, the high computational cost of adversarial training restricts their applicability in resource-constrained environments.

\subsection{High Dynamic Range Imaging}
High Dynamic Range (HDR) imaging aims to overcome the dynamic range limitations of a single exposure by computationally combining multiple images. The technical approaches to achieve this goal can be broadly categorized into two main paths: radiance map-based reconstruction and fusion-based reconstruction.

Radiance map-based methods first estimate a high-bit-depth radiance map from multiple LDR images, representing the real-world scene radiance. A tone-mapping operator then processes this map to compress its dynamic range for display on standard devices. While this path can provide high-fidelity lighting information, it is often more computationally expensive.

In contrast, fusion-based methods, such as the Multi-Exposure Fusion (MEF) techniques discussed in the previous section, directly combine multiple LDR images to generate a final image with a visually high dynamic range. By forgoing the intermediate radiance map estimation, these methods are computationally more efficient, making them particularly suitable for real-time applications on edge devices. This study focuses on the fusion-based technical path, aiming to develop a framework that achieves high efficiency and high-quality HDR reconstruction.

Deep learning techniques have been applied not only to MEF but have also been extended to the broader field of HDR image processing. DeepHDR~\cite{wu2017} proposed a non-flow-based deep learning framework capable of generalizing to different reference images while significantly reducing color artifacts and geometric distortions. ExpandNet~\cite{marnerides2018} introduced a multi-scale CNN architecture that learns  local and global information through separate branches to improve image quality. AHDRNet~\cite{yan2019} employed an attention-guided network to suppress irrelevant regions, combined with Dilated Residual Dense Blocks (DRDBs) to reconstruct missing details, effectively generating ghost-free HDR images.

GAN-based approaches have also been applied to HDR imaging. HDR-GAN~\cite{niu2021} was the first GAN-based method for HDR reconstruction, generating realistic content in regions with missing information through adversarial learning. It introduced a Reference-based Residual Merging module to align significant object motion in the feature domain and adopted deep HDR supervision to reduce artifacts during HDR reconstruction. UPHDR-GAN~\cite{li2022} proposed a multi-exposure HDR fusion network that can be trained on unpaired datasets while producing HDR results with fewer ghosting artifacts and defects. Despite their promise, these GAN-based methods face challenges, including high computational demand and dataset limitations. Accordingly, our study focuses on reducing computational burden and resource requirements while enhancing detail preservation and visual realism.

To further address ghosting in HDR images,~\cite{liu2022} proposed a Context-Aware Vision Transformer (CA-ViT), which captures both global and local dependencies so that local and global contexts operate in a complementary manner. Compared to conventional CNNs, CA-ViT more effectively mitigates ghosting caused by significant object motion. However, it remains computationally intensive, particularly for high-resolution images, which can become a bottleneck. In this study, we draw inspiration from this method and aim to refine it to reduce computational complexity while maintaining high performance and enabling deployment on resource-constrained edge devices.

\section{Methodology}

\begin{figure*}[t]
    \centering
    \includegraphics[width = \linewidth]{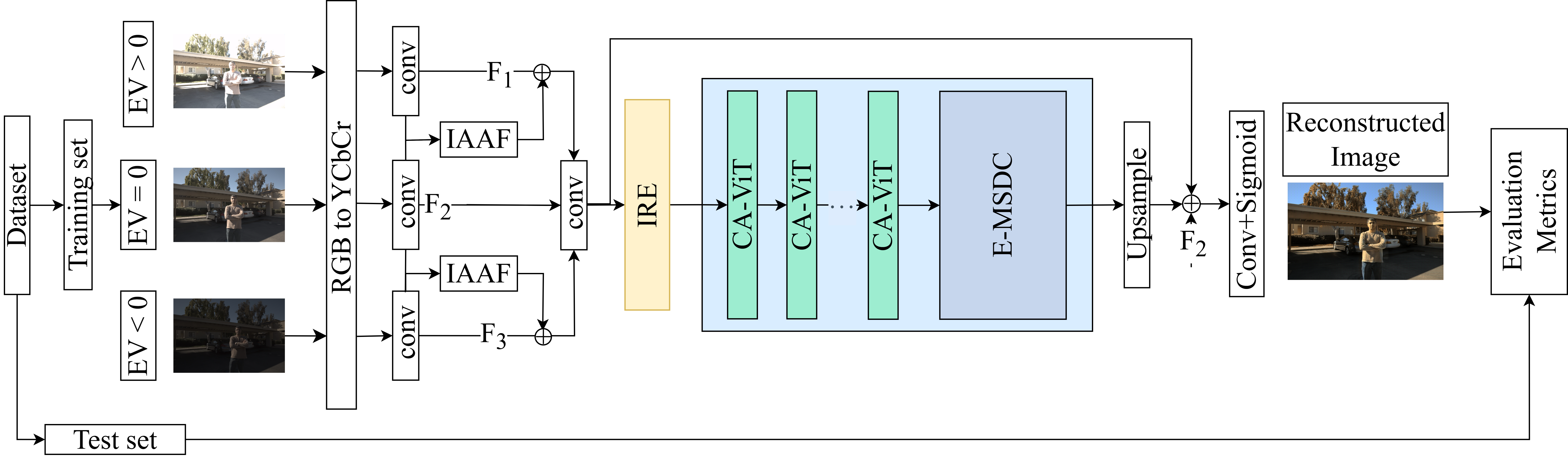}
    \caption{The overall architecture of EfficienT-HDR. The framework fuses three LDR images captured at different Exposure Values (EVs): underexposed (EV $<$ 0), normally exposed (EV $=$ 0), and overexposed (EV $>$ 0).}
    \label{fig:System_Architecture}
\end{figure*}

\subsection{Overall System Architecture}
This study builds upon the Context-Aware Vision Transformer (CA-ViT)~\cite{liu2022} and proposes a novel light-weight architecture, as illustrated in Fig.~\ref{fig:System_Architecture}. The framework can be divided into four main components. First, multiple RGB input images with different exposures are converted into the YCbCr color space to separate luminance and chrominance information, enabling the model to more effectively focus on processing and fusing luminance details across varying exposures. Second, an Intersection-Aware Adaptive Fusion (IAAF) module~\cite{weng2024}\!\cite{weng2024rethinking} is used to learn both the intersections and differences between features, generating fused feature maps while significantly reducing computational complexity. Third, conventional Patch Embedding is replaced with an innovative Inverted Residual Embedding (IRE)~\cite{sandler2019}, which is combined with stacked CA-ViT modules. These modules perform deep feature learning through two key internal components: Multi-head Self-Attention (MSA), which captures global, long-range contextual information, and a Local Context Extractor (LCE), which efficiently models local neighborhood features using convolutions. This dual approach ensures the model learns both broad and fine-grained details. To further optimize the architecture, Dynamic Tanh (DyT)~\cite{zhu2025} replaces the standard Layer Normalization in the Transformer, and our Enhanced Multi-Scale Dilated Convolution (E-MSDC)~\cite{gao2025} further captures multi-scale features. Finally, the processed features are fused and reconstructed to generate the final output image.

\begin{figure}[t]
    \centering
    \begin{subfigure}[b]{0.48\linewidth}
        \centering
        \includegraphics[width = \linewidth]{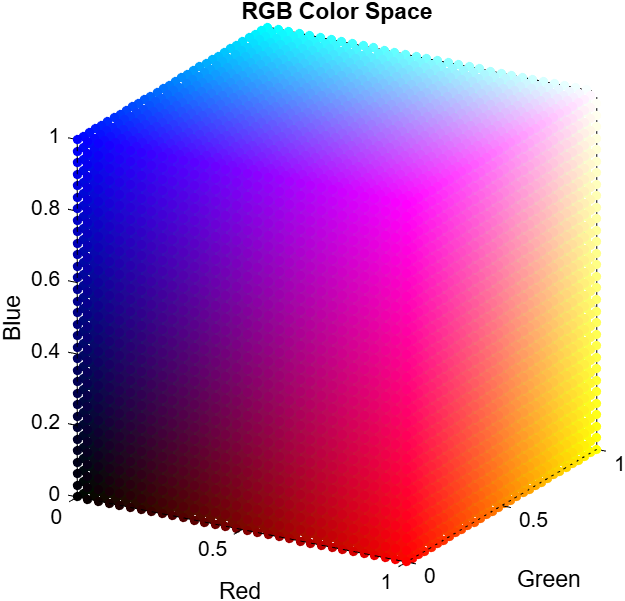}
        \caption{RGB Color Space Model}
        \label{fig:rgb_space_model}
    \end{subfigure}
    \hfill
    \begin{subfigure}[b]{0.48\linewidth}
        \centering
        \includegraphics[width = \linewidth]{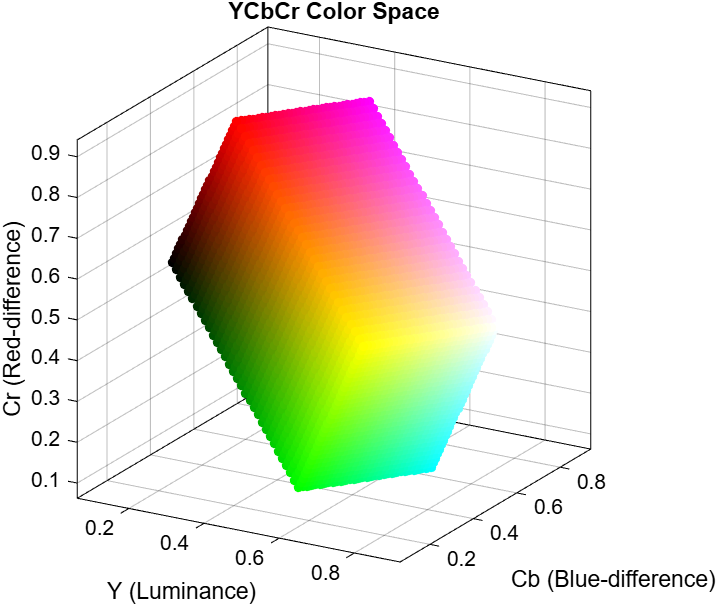}
        \caption{YCbCr Color Space Model}
        \label{fig:ycbcr_space_model}
    \end{subfigure}
    \caption{Visualization of the RGB and YCbCr color spaces. (a) The RGB model is an additive color space based on three orthogonal axes: Red, Green, and Blue. (b) The YCbCr model is a transformed space that decouples color information into a luminance (Y) axis and two chrominance (Cb, Cr) axes. This structural difference is key to our approach.}
    \label{fig:color_space_models}
\end{figure}

\subsection{Feature Extraction Based on Prior Knowledge}
In multi-exposure image fusion tasks, the traditional RGB color space couples luminance and chrominance information, which hinders the model's ability to perform targeted learning. To address this, this study introduces the concept of Channel Prior at the very front of the feature extraction stage by converting all input RGB images into the YCbCr color space. The fundamental structural differences between these two color spaces are visualized in Fig.~\ref{fig:color_space_models}.

The YCbCr color space is widely adopted as an effective color representation. As noted by Kolkur et al.~\cite{kolkur2016}, the core idea of YCbCr is to perform a nonlinear encoding of the conventional RGB space, thus decomposing image information into three key components: luminance (Y), chrominance-Blue ($C_b$), and chrominance-Red ($C_r$). This transformation is illustrated by the geometric difference between the standard additive cube of the RGB model (Fig.~\ref{fig:rgb_space_model}) and the resulting parallelepiped of the YCbCr space (Fig.~\ref{fig:ycbcr_space_model}). The corresponding conversion formulas are as follows:
\begin{align}
    Y & = 0.299R + 0.287G + 0.11B \\
    Cb & = B - Y \\
    Cr & = R - Y 
\end{align}
This separation facilitates data compression and reduces computational overhead, enabling the model to focus more effectively on learning features across different dimensions. Moreover, due to the relative simplicity of this transformation, incorporating this prior knowledge allows the model to process multi-exposure images more efficiently, making it well-suited for deployment in resource-constrained environments, such as edge devices.

\subsection{Light-Weight Design}
This study adopts the Context-Aware Vision Transformer as the backbone. It implements targeted light-weight design strategies at multiple levels of the overall architecture, as illustrated in Fig.~\ref{fig:System_Architecture}. The goal is to significantly reduce the computational complexity and the number of model parameters while minimizing performance degradation.

\begin{figure*}[t]
    \centering
    \begin{subfigure}[t]{0.32\linewidth}
        \centering
        \includegraphics[width = \linewidth, height = 4.5cm, keepaspectratio]{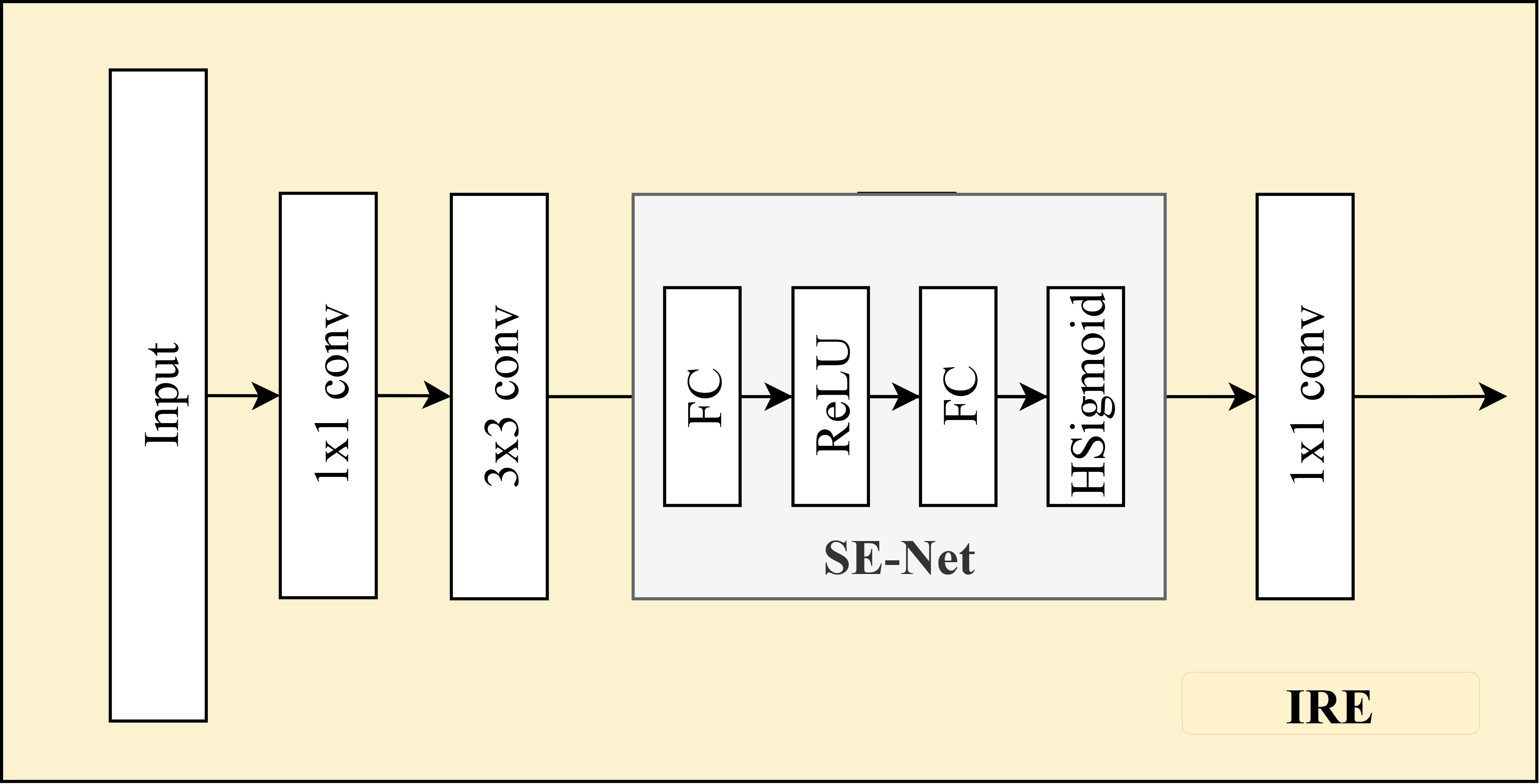}
        \caption{Inverted Residual Embedding (IRE)}
        \label{fig:ire_module}
    \end{subfigure}
    \hfill
    \begin{subfigure}[t]{0.32\linewidth}
        \centering
        \includegraphics[width = \linewidth, height = 4.5cm, keepaspectratio]{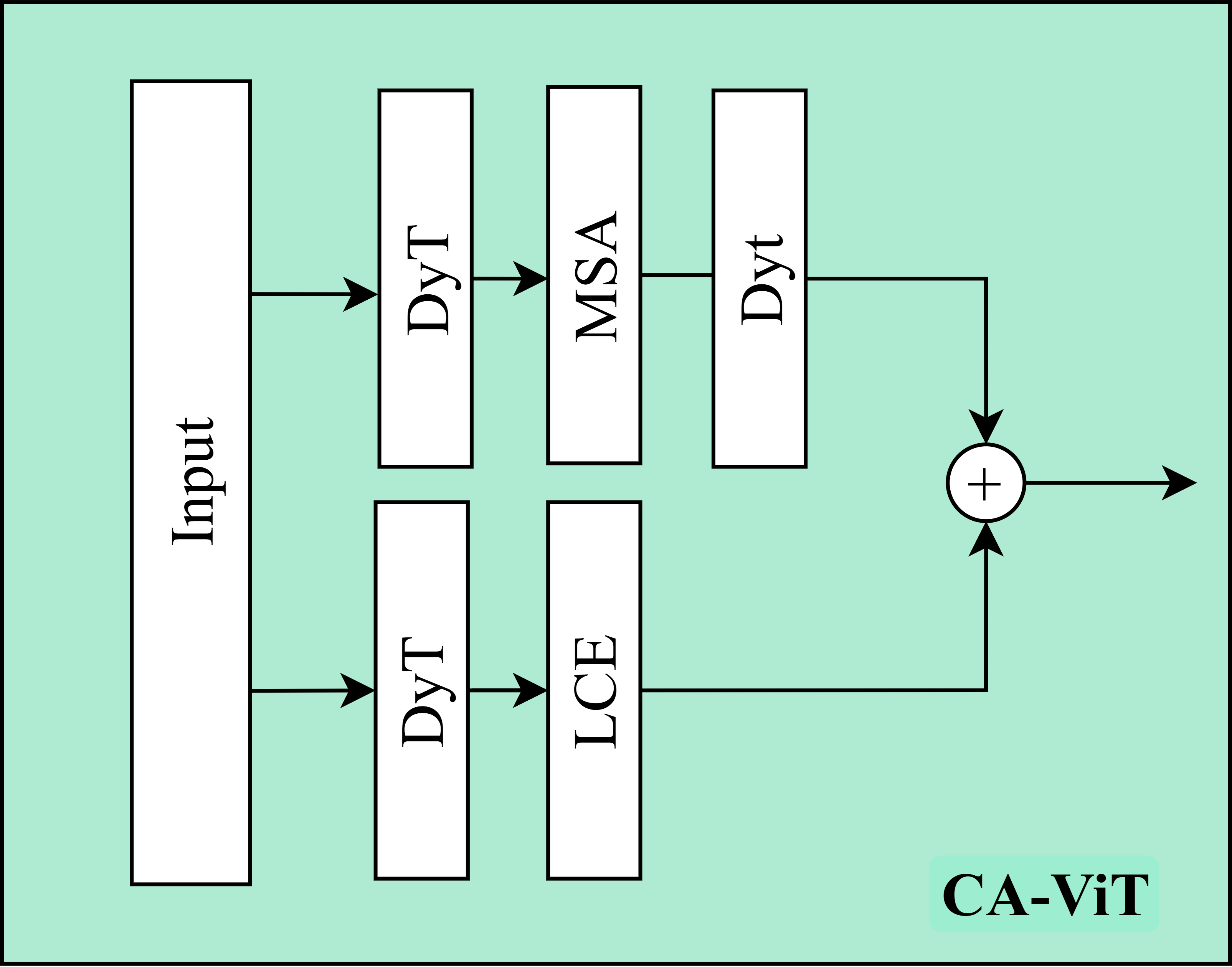}
        \caption{Context-Aware ViT (CA-ViT) Block}
        \label{fig:cavit_module}
    \end{subfigure}
    \hfill
    \begin{subfigure}[t]{0.32\linewidth}
        \centering
        \includegraphics[width = \linewidth, height = 4.5cm, keepaspectratio]{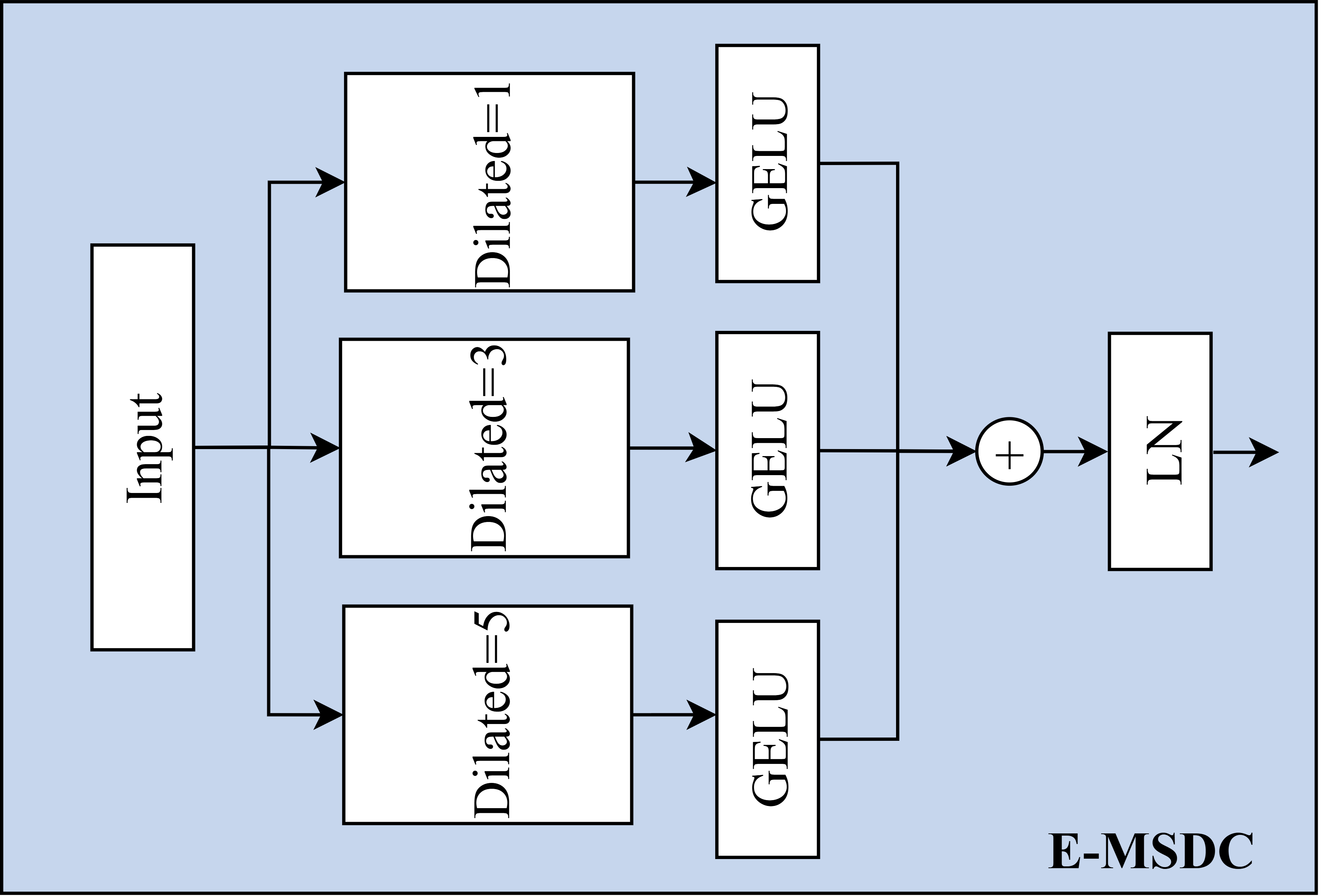}
        \caption{Enhanced Multi-Scale Dilated Convolution (E-MSDC)}
        \label{fig:emsdc_module}
    \end{subfigure}
    \caption{Architectures of the key modules proposed for our light-weight design. (a) The IRE module is used for efficient front-end feature extraction. (b) The CA-ViT block models global and local context. (c) The E-MSDC module provides low-cost multi-scale feature aggregation.}
    \label{fig:key_modules}
\end{figure*}

\subsubsection{Inverted Residual Embedding (IRE)}
A core challenge of the traditional Vision Transformer (ViT) architecture lies in the computational complexity of its self-attention mechanism. The computation grows quadratically with the number of input image patches, resulting in significant computational cost and memory usage for high-resolution image processing tasks. This complexity can be expressed as:
\begin{equation}
\Omega(\text{MSA}) = 4hwC^2 + 2(hw)^2 C
\label{eq:MSA_complexity}
\end{equation}
where $h$ and $w$ denote the number of image patches along height and width, respectively, and $C$ represents the feature dimension of each patch. To allow light-weight design without significant information loss, Zhao and Sun~\cite{zhao2025} proposed reducing the number of patches fed into the Transformer backbone, effectively mitigating computational cost and information degradation. 

Motivated by this, the present study introduces an Inverted Residual Embedding (IRE) module at the network frontend, replacing the conventional embedding layer. The IRE module is inspired by the core building block of the light-weight MobileNetV2 network~\cite{sandler2019}, aiming to serve as an efficient front-end backbone that extracts and compresses features before input into the Transformer. As illustrated in Fig.~\ref{fig:ire_module}, the internal structure of IRE follows an expansion–depthwise convolution–projection workflow and integrates a channel attention mechanism (SE-Net) to enhance feature representation. The operation of the module can be summarized as follows:
\begin{equation}
y = \text{Conv}_{1 \times 1} \Big( \text{SE}\big( \text{DWConv}_{3 \times 3, s}(\text{Conv}_{1 \times 1}(x)) \big) \Big)
\label{eq:IRE_operation}
\end{equation}
The key light-weight step occurs in the depthwise convolution stage, where a stride greater than 1 is applied to downsample the feature map. This reduces the number of patches entering the self-attention layer, fundamentally lowering the computational burden. Moreover, this convolution-based design introduces the inductive bias of CNNs into the model, strengthening the learning of local features. By introducing IRE, the embedding layer evolves from a simple data dimension projection into a strategic module that combines feature extraction and complexity control. This sets a solid foundation for the light-weight design of the overall network while maintaining high performance.

\subsubsection{Dynamic Tanh (DyT) Module}
Layer Normalization (LN) has long been regarded as indispensable for stabilizing Transformer training. However, LN necessitates computing per-feature statistics (mean and variance) at every forward pass, which increases computational overhead and conflicts with the goal of light-weight design. To this end, we adopt the Dynamic Tanh (DyT) module~\cite{zhu2025} as an alternative to LN. Empirically, the input–output mapping of trained LN layers exhibits an S-shaped curve akin to the hyperbolic tangent, which effectively rescales activations and compresses outliers. The core formulation is:
\begin{equation}
\mathrm{DyT}(x) = \gamma \cdot \tanh(\alpha x) + \beta
\label{eq:dyt}
\end{equation}
where $\alpha$ is a learnable scalar that dynamically scales the input features, and $\gamma$ and $\beta$ are learnable scale and shift vectors, respectively, analogous to those in LN. By leveraging the saturation property of 
$\tanh$, DyT emulates the nonlinear compression of LN while avoiding the computation of feature statistics, thereby reducing computational burden and better aligning with light-weight design.

\subsubsection{Enhanced Multi-Scale Dilated Convolution (E-MSDC) Module}

While Transformers can effectively establish global dependencies via self-attention, their ability to model multi-scale features within a single patch is relatively limited. Gao et al.~\cite{gao2025} observed that early-stage feature maps in their network retain high spatial resolution, leading to massive tokens; applying self-attention at this stage would incur prohibitive computational costs, becoming a significant bottleneck for light-weight design.

We introduce a light-weight yet effective token mixer named the Enhanced Multi-Scale Dilated Convolution (E-MSDC) module to address this issue. The design is inspired by the Multi-Scale Grouped Dilated Convolution (MSGDC) proposed by Gao et al.~\cite{gao2025} in their work on binarized networks. Still, we introduce critical adaptations to tailor it for our full-precision HDR reconstruction task.

As illustrated in Fig.~\ref{fig:emsdc_module}, the architecture of the E-MSDC module consists of multiple parallel grouped convolution branches. Each branch employs a 3x3 grouped convolution (\texttt{groups = 4}) but with a different dilation rate---specifically, rates of 1, 3, and 5 are used. This multi-branch, multi-dilation structure serves a dual purpose: the grouped convolutions significantly reduce the number of parameters and FLOPS. At the same time, the varying dilation rates provide receptive fields of different sizes. This allows the module to aggregate multi-scale contextual information and enhance the feature representation efficiently. For our HDR reconstruction task, we made two key modifications to the original design: first, the \texttt{RPReLU} activation function, which was intended for binarized models, is replaced with the smoother \texttt{GELU} activation function to better preserve the fine details essential for high-fidelity image restoration. Second, the entire E-MSDC module is innovatively integrated into the Context-Aware Transformer block's architecture.

This integration strategy is crucial to our model's performance. The E-MSDC module operates in a parallel path to the main self-attention and MLP layers within each Transformer block. Its output is then fused with the original input tensor $x$ via element-wise addition ($\text{output} = \text{E-MSDC}(\text{features}) + x$). This design establishes a complementary relationship: the self-attention mechanism focuses on capturing long-range, global dependencies, while the E-MSDC efficiently extracts rich, multi-scale local context. By fusing these features through a shortcut connection, the E-MSDC enhances the block's feature learning capability without incurring the quadratic complexity of additional attention layers.

\subsubsection{Intersection-Aware Adaptive Fusion (IAAF)}\label{sec:iaaf}

\begin{figure*}[t]
    \centering
    \includegraphics[width = 0.7\linewidth]{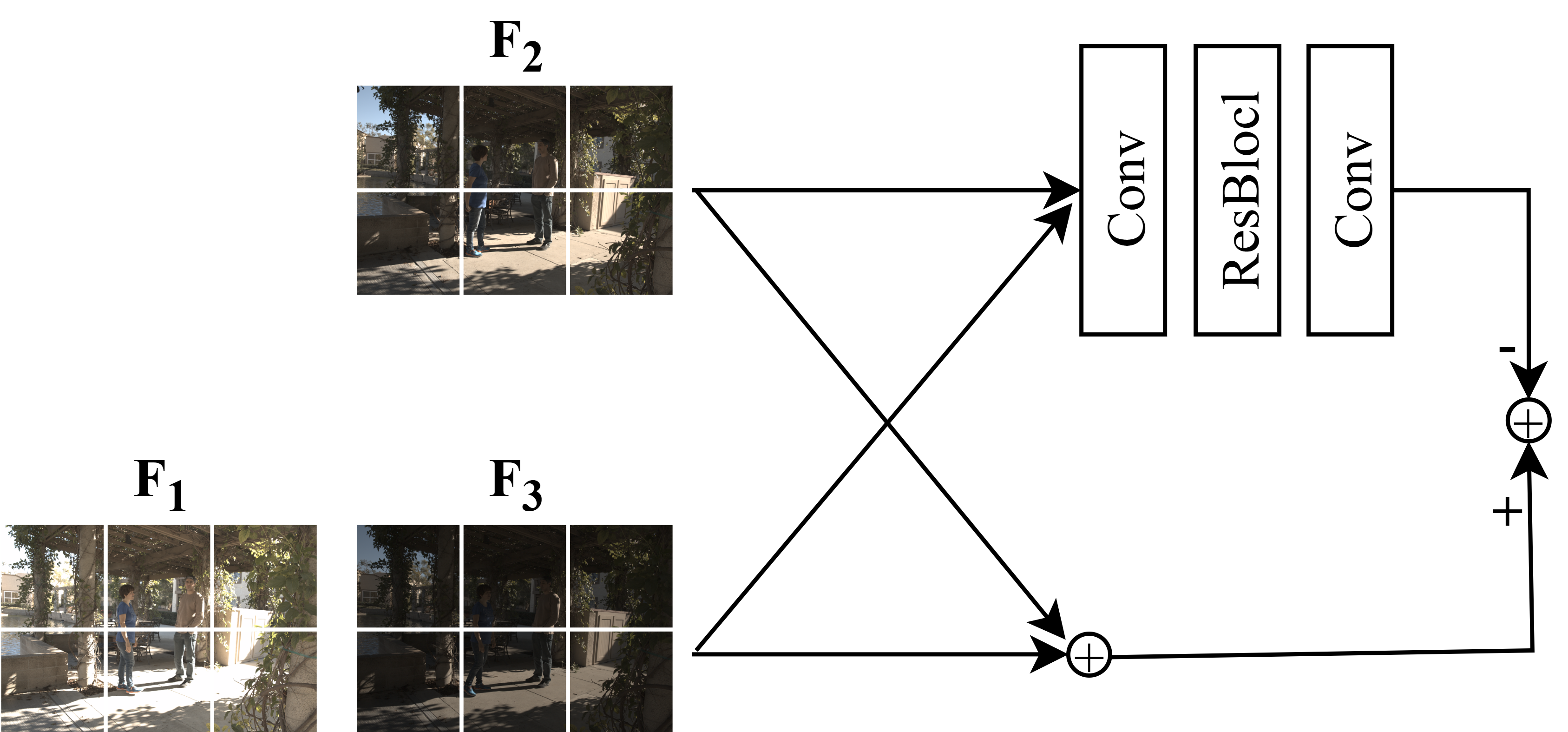}
    \caption{Architecture of IAAF.}
    \label{fig:IAAF}
\end{figure*}

In multi-exposure fusion under dynamic scenes, misalignment caused by object motion or camera shake is a primary source of ghosting artifacts. To address this, we replace the spatial attention in the baseline with the IAAF module~\cite{weng2024 }\!\cite{weng2024rethinking} . Rather than merely reweighting features, IAAF treats two exposure-specific feature maps $F_1$ and $F_2$ as sets and learns their shared information (intersection) via a light-weight convolutional network.

As illustrated in Fig.~\ref{fig:IAAF}, inputs $F_1,F_2,$ and $F_3$ correspond to over-, normal-, and under-exposed branches, respectively. For each pair, the two feature maps are concatenated and passed through a compact CNN composed of a convolution layer, a residual block, and another convolution layer to estimate the shared component. The fusion then follows the intersection-aware rule:
\begin{align}
F_{fused_1} &\approx F_1 + F_2 - \text{Intersection}(F_1,F_2) \\
F_{fused_3} &\approx F_3 + F_2 - \text{Intersection}(F_3,F_2)
\end{align}
This operation adds the two original feature maps and then subtracts the learned intersection produced by the convolutional network. By explicitly estimating and removing the common component, the model can focus on complementary, exposure-specific information, thereby preserving clear details from different exposures while suppressing redundancy or artifacts caused by misalignment. Because this convolution-based intersection computation scales linearly with the feature-map size, unlike spatial attention, whose matrix operations scale quadratically with the number of pixels, achieves feature alignment with fewer parameters and FLOPS, realizing a light-weight design.

\subsection{Analysis and Evaluation Metrics}

During model validation, we conduct three categories of performance assessment. First, we evaluate computational efficiency using FLOPS, inference time, and latency. Second, we assess model compactness by reporting the number of parameters to verify compliance with light-weight design goals. Finally, we measure image quality using PSNR (Peak Signal-to-Noise Ratio) and SSIM (Structural Similarity Index), each computed in two domains: \textbf{PSNR-$l$}, \textbf{PSNR-$\mu$}, \textbf{SSIM-$l$}, and \textbf{SSIM-$\mu$}. PSNR quantifies the error between two images, as in Eq.~(\ref{eq:psnr}), while SSIM evaluates similarity in terms of luminance, contrast, and structure, as in Eq.~(\ref{eq:ssim}):

\begin{equation}
\mathrm{PSNR} = 10 \log_{10}\!\left(\frac{\mathrm{MAX}^2}{\mathrm{MSE}}\right),
\label{eq:psnr}
\end{equation}
where $\mathrm{MSE}$ is the mean squared error between the fused image and the reference image, 
and $\mathrm{MAX}$ is the maximum possible pixel value.

\begin{equation}
\mathrm{SSIM} = l^{\alpha} \cdot c^{\beta} \cdot s^{\gamma},
\label{eq:ssim}
\end{equation}
where $l$, $c$, and $s$ denote the luminance, contrast, and structure components, respectively, 
and $\alpha$, $\beta$, and $\gamma$ are their corresponding weights.

\textbf{PSNR-$l$} computes the error between the fused and the reference HDR images in the original linear domain, probing whether actual luminance is preserved. 
\textbf{PSNR-$\mu$} performs the comparison in the $\mu$-law compressed domain, which is more aligned with human visual perception. 
Analogously, \textbf{SSIM-$l$} measures structural, luminance, and contrast similarity in the linear domain, while \textbf{SSIM-$\mu$} evaluates structural similarity in the $\mu$-law compressed domain, again better reflecting perceptual quality. 
Higher PSNR and SSIM values indicate closer agreement with the reference image and, consequently, better perceived quality.

Beyond the above metrics, to more accurately assess the perceptual quality of high dynamic range images, we additionally adopt \textbf{HDR-VDP-2} (High Dynamic Range Visible Difference Predictor)\!\cite{mantiuk2011hdrvdp2} as a key evaluation tool. 
Unlike PSNR and SSIM, HDR-VDP-2 is grounded in the human visual system (HVS) and more faithfully reflects human perception of image quality. 
It accounts for factors such as luminance adaptation, contrast sensitivity, and visual masking, and produces a perceptual quality score---higher values indicate closer agreement with human visual experience.

\section{Experiment Results}
\subsection{Implementation Details and Datasets}

\begin{figure}[t]
  \centering
  \begin{subfigure}[b]{0.32\linewidth}
    \centering
    \includegraphics[width = \linewidth]{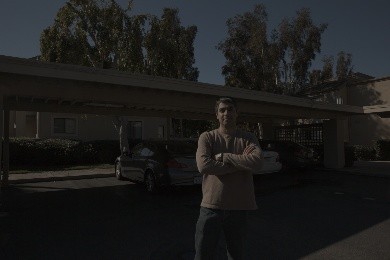}
    \caption{}
    \label{fig:low}
  \end{subfigure}\hfill
  \begin{subfigure}[b]{0.32\linewidth}
    \centering
    \includegraphics[width = \linewidth]{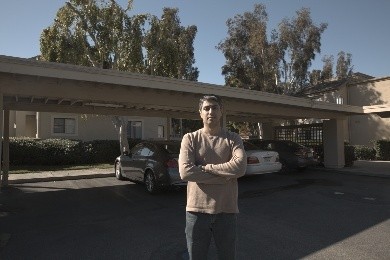}
    \caption{}
    \label{fig:normal}
  \end{subfigure}\hfill
  \begin{subfigure}[b]{0.32\linewidth}
    \centering
    \includegraphics[width = \linewidth]{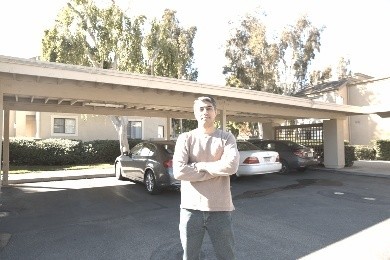}
    \caption{}
    \label{fig:high}
  \end{subfigure}
  \caption{Kalantari and  Ramamoorthi's  dataset~\cite{kalantari2017}. 
  (a) Under exposure; (b) Normal; (c) Over exposure.}
  \label{fig:kalantari_example}
\end{figure}

\subsubsection{Implementation Details}
All experiments were conducted using PyTorch. We adopted the AdamW optimizer with an initial learning rate of $1\times 10^{-4}$ and weight decay of $2\times 10^{-2}$. To promote stable convergence, the learning rate was scheduled using Cosine Annealing Warm Restarts with an initial restart period $T_0 = 25$ epochs, period multiplier $T_{\text{mult}} = 1$, and minimum learning rate $\eta_{\min} = 1\times 10^{-9}$. Training employed a Joint Recon Perceptual Loss as the objective. The models were trained for 100 epochs with a batch size of 14. Input images were cropped into $128\times 128$ patches for training.

\subsubsection{Datasets}
We train on the Kalantari and Ramamoorthi's dataset~\cite{kalantari2017}. The data consist of 1500$\times$1000 exposure-bracketed sequences with 74 groups for training and 10 for testing. Each sequence contains three LDR images captured at different exposure values and one high-quality HDR reference for supervision and evaluation. The dataset covers diverse dynamic scenarios, including city streets, crowds, and moving vehicles, as shown in Fig.\ref{fig:kalantari_example}.

\subsection{Evaluation of the Image Fusion Models}

\begin{table*}[t]
    \centering
    \caption{Quantitative comparison on the Kalantari and Ramamoorthi's dataset~\cite{kalantari2017} dataset. We represent the first and second ranks with \textbf{bold} and \underline{underlined}, respectively, and the third with \emph{italics}}
    \label{tab:kalantari_results}
    \setlength{\tabcolsep}{6pt}
    \begin{tabular}{lccccc}
    \hline
    Model & PSNR-$l$ $\uparrow$ & SSIM-$l$ $\uparrow$ & PSNR-$\mu$ $\uparrow$ & SSIM-$\mu$ $\uparrow$ & HDR-VDP-2 $\uparrow$ \\
    \hline
    HDR-Transformer~\cite{liu2022}   & \textbf{39.252} & \textbf{0.9885} & \textbf{42.719} & \textbf{0.9919} & \emph{64.63} \\
    Sen12~\cite{sen2012}             & 38.110          & 0.9721          & 40.800          & 0.9808          & 59.38        \\
    Kalantari17~\cite{kalantari2017} & 38.158          & 0.9775          & 38.737          & 0.9807          & 63.21        \\
    DeepHDR~\cite{wu2017}            & 32.703          & 0.9002          & 31.058          & 0.8499          & 59.39        \\
    \hline
    Ours                             & \underline{38.539} & \underline{0.9856} & \underline{41.487} & \underline{0.9891} & \textbf{65.09} \\
    Ours-Lite                        & \emph{38.200}   & \emph{0.9854}   & \emph{41.060} & \emph{0.9888}   & \underline{64.88} \\
    \hline
    \end{tabular}
\end{table*}

We compare our main model (Ours) and the light-weight variant (Ours-Lite) against state-of-the-art multi-exposure HDR methods on the Kalantari and  Ramamoorthi's dataset. The quantitative results are summarized in Table \ref{tab:kalantari_results}. As the data show, both  variants deliver competitive performance across all metrics. In direct comparison with the original HDR-Transformer, our models achieve comparable PSNR and SSIM while  attaining superior perceptual quality on HDR-VDP-2. In general, the results indicate that both versions produce high-quality HDR outputs. Despite minor differences in traditional metrics, their advantage on the perceptual indicator (HDR-VDP-2) highlights the strength of the proposed architecture in visual fidelity and preservation of detail.

\subsection{Visual Analysis of Fusion Results}

\begin{figure*}[t]
    \centering
    \includegraphics[width = 0.85\linewidth]{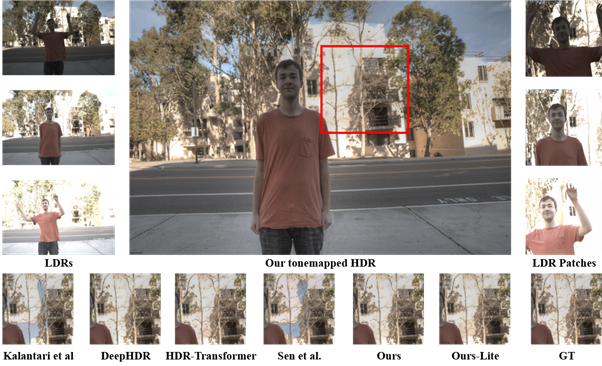}
    \caption{Visualization results of different methods on the Kalantari and Ramamoorthi's dataset.}
    \label{fig:test_results}
\end{figure*}

Fig.\ref{fig:test_results} visually compares a Kalantari and  Ramamoorthi's dataset with significant foreground motion and a complex background. All HDR results are rendered with the same tone-mapping operator before display to ensure consistency and fairness. For fine-grained inspection, the red boxed region at the image center is magnified and shown beneath each result. As observed, several methods struggle in this scene; for example, Sen12~\cite{sen2012} and Kalantari17~\cite{kalantari2017} exhibit pronounced ghosting artifacts.

In contrast, our main model (Ours) produces sharp object contours, effectively suppresses ghosting, and restores the structural details of the background buildings, yielding a visual appearance closest to the ground truth (GT). Nevertheless, a slight overexposure can be noticed in high-frequency details (e.g., tree branches), with a minor loss of local contrast. The light-weight variant (Ours-Lite) also suppresses most ghosting. Still, it shows somewhat reduced edge sharpness around moving objects and weaker separation from the background compared to the main model, reflecting the trade-off made to maximize computational efficiency.

In general, this qualitative comparison substantiates the robustness of the proposed models in complex dynamic scenes: the main version achieves state-of-the-art ghost suppression. In contrast, the light-weight version achieves high efficiency with acceptable visual quality.

\subsection{Model Efficiency and Light-Weight Analysis}

\begin{table*}[t]
    \centering
    \caption{Quantitative evaluation of computational complexity and inference speed on the Kalantari and Ramamoorthi's dataset. We represent the first and second ranks with \textbf{bold} and \underline{underlined}, respectively, and the third with \emph{italics}}
    \label{tab:complexity}
    \setlength{\tabcolsep}{6pt}
    \begin{tabular}{lcccc}
    \hline
    Model & FLOPS (G) $\downarrow$ & Parameters (M) $\downarrow$ & GPU (ms/image) $\downarrow$ & CPU (ms/patch) $\downarrow$ \\
    \hline
    HDR-Transformer~\cite{liu2022}   & 21.61               & 1.22                & 94.43               & 662.68 \\
    Kalantari17~\cite{kalantari2017} & 12.57               & \textbf{0.39}       & \textbf{2.81}       & \textbf{28.85} \\
    DeepHDR~\cite{wu2017}            & \emph{7.11}         & 16.61               & \underline{6.32}    & -- \\
    \hline
    Ours                             & \underline{7.04}    & \emph{1.14}         & 66.55               & \emph{125.81} \\
    Ours-Lite                        & \textbf{5.98}       & \underline{1.07}    & \emph{66.26}        & \underline{116.06} \\
    \hline
    \end{tabular}
\end{table*}

This section provides a quantitative analysis of computational efficiency and resource demands to verify whether the proposed architecture achieves its light-weight design objectives and to assess its potential for deployment on edge devices. 

We test the inference speed on both GPU and CPU to evaluate practical runtime performance. The GPU tests are conducted on an NVIDIA RTX 3050 Ti, reporting the average time over 100 forward passes for a single image. The CPU tests run on an Intel Core i7-11800H, where models are exported to ONNX and measured per image patch. The detailed comparison results are summarized in Table ~\ref{tab:complexity}.

Although our two variants do not attain the best scores on all image quality metrics, they deliver substantial gains in computational complexity and model size. Compared to the reference HDR-Transformer, the main model (Ours) reduces the FLOPS from 21.61 G to 7.04 G ($\approx67\%$ reduction). Owing to the  reduced computation, the ONNX CPU runtime is over 5$\times$ faster, while the GPU inference speed is 30\% faster.

The light-weight variant (Ours-Lite) pushes efficiency further: its FLOPS and parameter count are only 5.98 G and 1.07 M, respectively, the most compact among all compared models. Crucially, Ours-Lite outperforms the main version across all key efficiency indicators (FLOPS, parameter count, GPU/CPU inference speed), underscoring its role as the high-efficiency option. The results show that both architectures achieve significant light-weighting while maintaining high-quality output. Their low computational cost, compact model size, and high throughput demonstrate strong potential for real-time processing on resource-constrained edge devices.

\subsection{Edge Device Deployment and Inference Speed Analysis}

\begin{table}[t]
\centering
\caption{Jetson NX per-patch inference time (ms) comparison. We represent the first rank in \textbf{bold} and the second in \underline{underlined}.}
\label{tab:jetson_nx}
\setlength{\tabcolsep}{10pt}
\begin{tabular}{lc}
\hline
Model & Jetson-NX (ms/patch) $\downarrow$ \\
\hline
HDR-Transformer & 2151.68 \\
Ours            & \underline{857.55} \\
Ours-Lite       & \textbf{800.47} \\
\hline
\end{tabular}
\end{table}

We deploy the models on a resource-constrained edge platform to assess their practical inference performance. The target hardware is the NVIDIA Jetson Xavier NX development kit, equipped with a 6-core ARMv8 CPU and an integrated NVIDIA Tegra Xavier GPU, which is considered more limited than desktop GPUs. Measuring latency on this platform offers a realistic estimate of runtime efficiency under edge constraints. We report the average per-patch inference time for the main model (Ours), the light-weight variant (Ours-Lite), and the HDR-Transformer baseline. Detailed results are shown in Table \ref{tab:jetson_nx}.

As shown, our models deliver substantial speed advantages on the edge device. The HDR-Transformer takes 2151.68 ms/patch, whereas Ours requires 857.55 ms/patch. Ours-Lite achieves 800.47 ms/patch, resulting in an approximately 2.69$\times$ speedup over the baseline. These results confirm that our light-weighting strategies translate into clear runtime gains on edge hardware, indicating strong potential for real-time deployment in resource-constrained applications.

\subsection{Ablation Study}
To validate the effectiveness of key design choices, we conduct ablations on: (1) the Inverted Residual Embedding (IRE); (2) the activation selection within E-MSDC; and (3) the differences between two models.

\subsubsection{Inverted Residual Embedding (IRE)}

\begin{table}[h!]
\centering
\caption{Image quality with and without Inverted Residual Embedding (IRE) (\textbf{bold} = best; \checkmark = with IRE).}
\label{tab:ablation_ire_quality}
\setlength{\tabcolsep}{4pt}
\begin{tabular}{lccccc}
\hline
IRE & PSNR-$l$ $\uparrow$ & SSIM-$l$ $\uparrow$ & PSNR-$\mu$ $\uparrow$ & SSIM-$\mu$ $\uparrow$ & HDR-VDP-2 $\uparrow$ \\
\hline
\checkmark & 38.0760 & 0.9867 & \textbf{40.9750} & 0.9891 & 64.55 \\
 & \textbf{38.7190} & \textbf{0.9869} & 40.7420 & \textbf{0.9897} & \textbf{65.19} \\
\hline
\end{tabular}
\end{table}

\begin{table}[h!]
\centering
\caption{Efficiency with and without Inverted Residual Embedding (IRE) (\textbf{bold} = best; \checkmark = with IRE).}
\label{tab:ablation_ire_eff}
\setlength{\tabcolsep}{6pt}
\begin{tabular}{lcc}
\hline
IRE & FLOPS (G) $\downarrow$ & Parameters (M) $\downarrow$ \\
\hline
\checkmark      & \textbf{12.10} & 1.45 \\
                & 20.95          & \textbf{1.19} \\
\hline
\end{tabular}
\end{table}

We compare the proposed frontend IRE with an ablated model that replaces IRE with a conventional patch embedding. Table \ref{tab:ablation_ire_quality} summarizes that the conventional embedding attains higher HDR-VDP-2, whereas IRE yields a small but consistent gain on PSNR-$\mu$. In terms of efficiency (Table \ref{tab:ablation_ire_eff}), IRE reduces FLOPS from 20.95 G to 12.10 G. Given our goal of achieving fast inference on edge devices, this trade-off—substantially reduced computation at the cost of a slight parameter increase—is highly favorable, highlighting the necessity of IRE for light-weight, efficient deployment.

\subsubsection{Choice of Activation in E-MSDC}

\begin{figure}[t]
  \centering
  \begin{subfigure}[b]{0.48\linewidth}
    \centering
    \includegraphics[width = \linewidth]{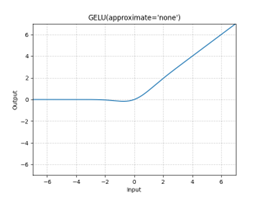}
    \caption{GELU}
    \label{fig:gelu}
  \end{subfigure}\hfill
  \begin{subfigure}[b]{0.48\linewidth}
    \centering
    \includegraphics[width = \linewidth]{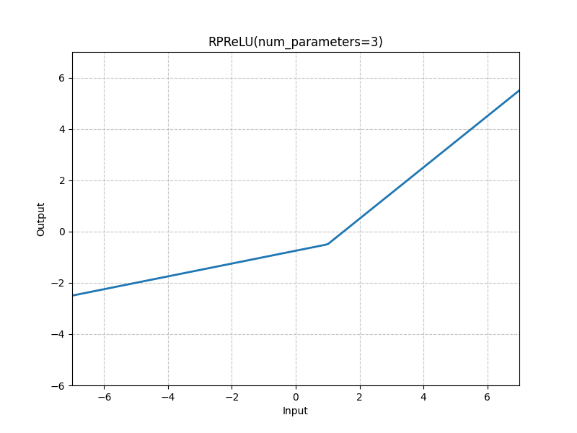}
    \caption{RPReLU}
    \label{fig:rprelu}
  \end{subfigure}
  \caption{Activation function curves.}
  \label{fig:curves}
\end{figure}

\begin{table}[h!]
\centering
\caption{Image quality comparison of activations in E-MSDC (\textbf{bold} = best).}
\label{tab:emsdc_quality}
\setlength{\tabcolsep}{4pt}
\begin{tabular}{lccccc}
\hline
$f(x)$ & PSNR-$l$ $\uparrow$ & SSIM-$l$ $\uparrow$ & PSNR-$\mu$ $\uparrow$ & SSIM-$\mu$ $\uparrow$ & HDR-VDP-2 $\uparrow$ \\
\hline
RPReLU & 37.8130 & 0.9860 & 39.7790 & 0.9866 & \textbf{65.27} \\
GELU   & \textbf{38.7190} & \textbf{0.9869} & \textbf{40.7420} & \textbf{0.9897} & 65.19 \\
\hline
\end{tabular}
\end{table}

\begin{table}
\centering
\caption{Efficiency with different activations in E-MSDC (\textbf{bold} = best).}
\label{tab:emsdc_eff}
\setlength{\tabcolsep}{6pt}
\begin{tabular}{lcc}
\hline
Activation & FLOPS (G) $\downarrow$ & Parameters (M) $\downarrow$ \\
\hline
RPReLU & \textbf{19.75} & 1.20 \\
GELU   & 20.95          & \textbf{1.19} \\
\hline
\end{tabular}
\end{table}

This ablation study validates our choice of activation function within the proposed E-MSDC module. The original MSGDC design by Gao et al.~\cite{gao2025} employed the \texttt{RPReLU} activation, which was primarily developed for binarized networks. We hypothesized that a smoother, nonlinear activation function would be preferable for full-precision HDR reconstruction—a task where fine detail recovery is critical. To verify this, we directly compared our E-MSDC module implemented with two different activation functions: the original \texttt{RPReLU} and the proposed \texttt{GELU}. The response curves of these functions are illustrated in Fig.~\ref{fig:curves}, with quantitative results for image quality and model efficiency reported in Tables \ref{tab:emsdc_quality} and \ref{tab:emsdc_eff}, respectively.

The experimental results clearly support our hypothesis. As shown in Table~\ref{tab:emsdc_quality}, the variant using \texttt{GELU} outperforms the \texttt{RPReLU} version across all four standard fidelity metrics (PSNR-$l$, SSIM-$l$, PSNR-$\mu$, and SSIM-$\mu$), indicating superior performance in preserving image structure and luminance. While the \texttt{RPReLU} version holds a marginal advantage in the perceptual HDR-VDP-2 score, the data in Table~\ref{tab:emsdc_eff} confirms that the choice of activation has a negligible impact on computational complexity and parameter count. Given that \texttt{GELU} provides a better overall balance and is more aligned with high-fidelity, full-precision image reconstruction requirements, we adopt it as the standard activation function in our final E-MSDC module.

\subsubsection{Architectural Analysis}

\begin{figure*}
    \centering
    \includegraphics[width = \linewidth]{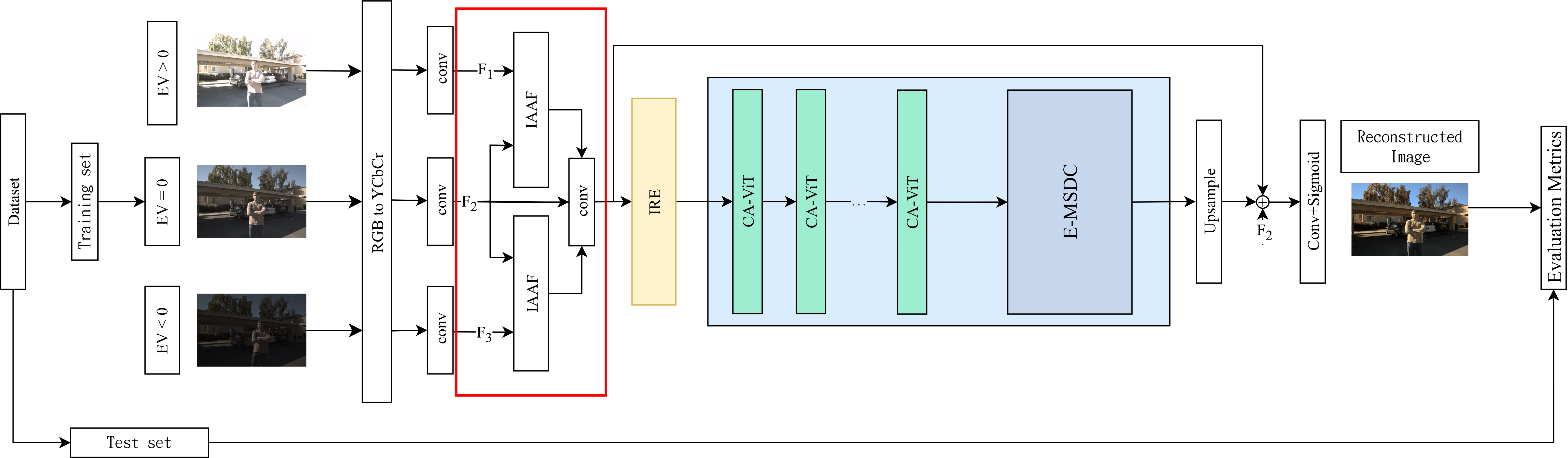}
    \caption{The architecture of our proposed method in light-weight version. The differences from the main model are highlighted with a red box.}
    \label{fig:lite_Architecture}
\end{figure*}

We provide two variants whose main architectural difference lies in the frontend fusion strategy, which directly affects computational load and information retention. Let 
$C$ denote the number of base feature channels extracted by the initial convolutional stem.
\begin{itemize}
    \item \textbf{Main model (Ours):} As in Fig.~\ref{fig:System_Architecture}, features processed by the IAAF module (introduced in Section \ref{sec:iaaf}) are concatenated with the original features to form two enriched feature blocks, which are then fused with the middle-frame features $F_2$. This yields a total input channel budget of $5C$ to subsequent stages.
    \item \textbf{light-weight model (Ours-Lite):} Trades some raw information for higher efficiency. As shown in Fig.~\ref{fig:lite_Architecture}, concatenating only the aligned features with $F_2$ yields $3C$ channels. This reduces downstream convolutional cost roughly proportional to input channels, thereby improving runtime with minimal quality loss.
\end{itemize}

Overall, the main variant pursues higher-fidelity image reconstruction through deeper feature interactions. In contrast, the light-weight variant streamlines the fusion pipeline to deliver superior computational efficiency while maintaining acceptable visual quality.

\section{Conclusion}

This study successfully addresses the critical challenge of achieving high-quality, ghost-free High Dynamic Range (HDR) imaging on resource-constrained edge devices. To this end, we designed and implemented EfficienT-HDR, a novel light-weight Vision Transformer framework that improves the existing fusion-based technical path. We proposed two model versions tailored to different application requirements: a main version that pursues ultimate image quality and a light-weight version focused on computational efficiency.

Comprehensive quantitative and qualitative evaluations demonstrate the superiority of the proposed architecture. The main version achieves state-of-the-art performance across all image quality metrics, surpassing existing methods, particularly excelling in the perceptual HDR-VDP-2 index. In terms of efficiency, compared with the baseline model, its FLOPS are substantially reduced by 67\%. At the same time, inference speeds on the CPU and the Jetson NX edge device are improved by more than fivefold and 2.5 times, respectively. Meanwhile, the light-weight version exhibits even greater efficiency across all key performance indicators, making it highly suitable for deployment in scenarios with minimal resources.

In summary, by systematically integrating multiple key light-weight technologies—including Inverted Residual Embedding (IRE)~\cite{sandler2019} for efficient feature extraction, the Intersection-Aware Adaptive Fusion (IAAF) module~\cite{weng2024}\!\cite{weng2024rethinking} for ghost suppression, the Dynamic Tanh (DyT) function~\cite{zhu2025}, and our proposed Enhanced Multi-Scale Dilated Convolution (E-MSDC)~\cite{gao2025}—this work not only achieves an excellent balance between performance and efficiency in a single model but also provides a flexible and powerful solution framework for advancing the adoption of high-quality HDR technology in diverse, real-time dynamic applications within edge computing environments.

\section*{Acknowledgements}
This research was supported by the National Science and
Technology Council (NSTC) of Taiwan, under the College
Student Research Scholarship, grant number 114-2813-C-033-
037-E.
\FloatBarrier

\bibliographystyle{IEEEtran}
\bibliography{ref}

\end{document}